\documentclass[lettersize,journal]{IEEEtran}
\usepackage{amsmath,amsfonts}
\usepackage{algorithmic}
\usepackage{array}
\usepackage[caption=false,font=normalsize,labelfont=sf,textfont=sf]{subfig}
\usepackage{textcomp}
\usepackage{stfloats}
\usepackage{url}
\usepackage{verbatim}
\usepackage{graphicx}
\usepackage{cite}
\usepackage{xspace}
\usepackage{xcolor}
\usepackage{booktabs}
\usepackage{hyperref}
\usepackage{threeparttable}
\usepackage{multirow}
\usepackage{amssymb}
\usepackage{wrapfig}
\usepackage[T1]{fontenc}
\def\BibTeX{{\rm B\kern-.05em{\sc i\kern-.025em b}\kern-.08em
    T\kern-.1667em\lower.7ex\hbox{E}\kern-.125emX}}
\usepackage{balance}

\newcommand{\method}{TinyRS\xspace}

\newif\ifrevision

\revisionfalse

\colorlet{revision}{black}
\ifrevision \colorlet{revision}{blue}\fi

\begin{document}

\title{\method-R1: Compact Vision Language Model for Remote Sensing}

\author{Aybora Köksal,~\IEEEmembership{Student Member,~IEEE,}
A. Aydın Alatan,~\IEEEmembership{Senior Member,~IEEE}
\vspace{-.6cm}
\thanks{Aybora Köksal and A. Aydın Alatan are with Center for the Image Analysis (OGAM) and Department of Electrical and Electronics Engineering of Middle East Technical University (METU), Ankara, Turkey (e-mail: aybora@metu.edu.tr, alatan@metu.edu.tr)}
\thanks{Manuscript received August 6, 2025; revised September 26, 2025\textcolor{revision}{; accepted October 15, 2025.}}}


\markboth{IEEE Geoscience and Remote Sensing Letters, 2025}{Köksal, Alatan: TinyRS-R1: Compact VLM for Remote Sensing}

\maketitle

\begin{abstract}

Remote sensing applications often rely on edge hardware that cannot host the models in the 7B parametric vision language of today. This paper presents \method, the first 2B-parameter VLM optimized for remote sensing, and \method-R1, its reasoning-augmented variant. Based on Qwen2-VL-2B, \method is trained via a four-stage pipeline: pre-training on million-scale satellite images, instruction tuning, fine-tuning with Chain-of-Thought (CoT) annotations from a new reasoning dataset, and GRPO-based alignment. \method-R1 matches or surpasses recent 7B remote sensing models in classification, VQA, grounding, and open-ended QA--while using one third of the memory and latency. CoT reasoning improves grounding and scene understanding, while \method excels at concise, low-latency VQA. \method-R1 is the first domain-specialized small VLM with GRPO-aligned CoT reasoning for general-purpose remote sensing. \textcolor{revision}{The code, models, and caption datasets are available at \url{https://github.com/aybora/TinyRS}}.

\end{abstract}


\vspace{-.4cm}
\section{Introduction}
\label{sec:intro}

Vision-language models (VLMs) like GPT-4V \cite{gpt4v} and \textcolor{revision}{open alternatives} such as Qwen2-VL \cite{wang2024qwen2} and InternVL2 \cite{chen2024internvl} have redefined natural language and visual understanding, setting new benchmarks in vision-language tasks. However, their general-purpose design and high computational demands limit \textcolor{revision}{applicability} in specialized or resource-constrained settings. These challenges have fueled interest in Small Language Models (SLMs, <3B parameters) \cite{liu2024mobilellmoptimizingsubbillionparameter} and Small Vision-Language Models (SVLMs) \cite{wang2024qwen2, chen2024internvl}. This shift has also driven \textcolor{revision}{development} of domain-specialized models for tasks requiring expert knowledge, where general-purpose models often fall short.

Remote sensing (RS) requires specialized models due to its distinct nature from natural imagery. Building on VLM progress in vision tasks, recent work has adapted them for RS, producing several 7B-parameter models for VQA, detection, and grounding \cite{hu2307remote, kuckreja2024geochat, muhtar2024lhrs, pang2025vhm}. These advances highlight the growing role of VLMs in the semantic and spatial understanding of RS. However, their large size makes them unsuitable for edge devices, not counted as SVLMs. Moreover, these models do not take advantage of reinforcement learning for Chain-of-Thought (CoT) alignment, with two concurrent exceptions: Muhtar et al. \cite{muhtar2025quality} use GRPO for answer quality but report modest gains, while MilChat \cite{koksal2025milchat} applies RL effectively but its focus is limited to a single RS task: military base classification.


\begin{figure*}
\vspace{-.8cm}
\begin{tabular}{ccc}
\includegraphics[width=\linewidth]{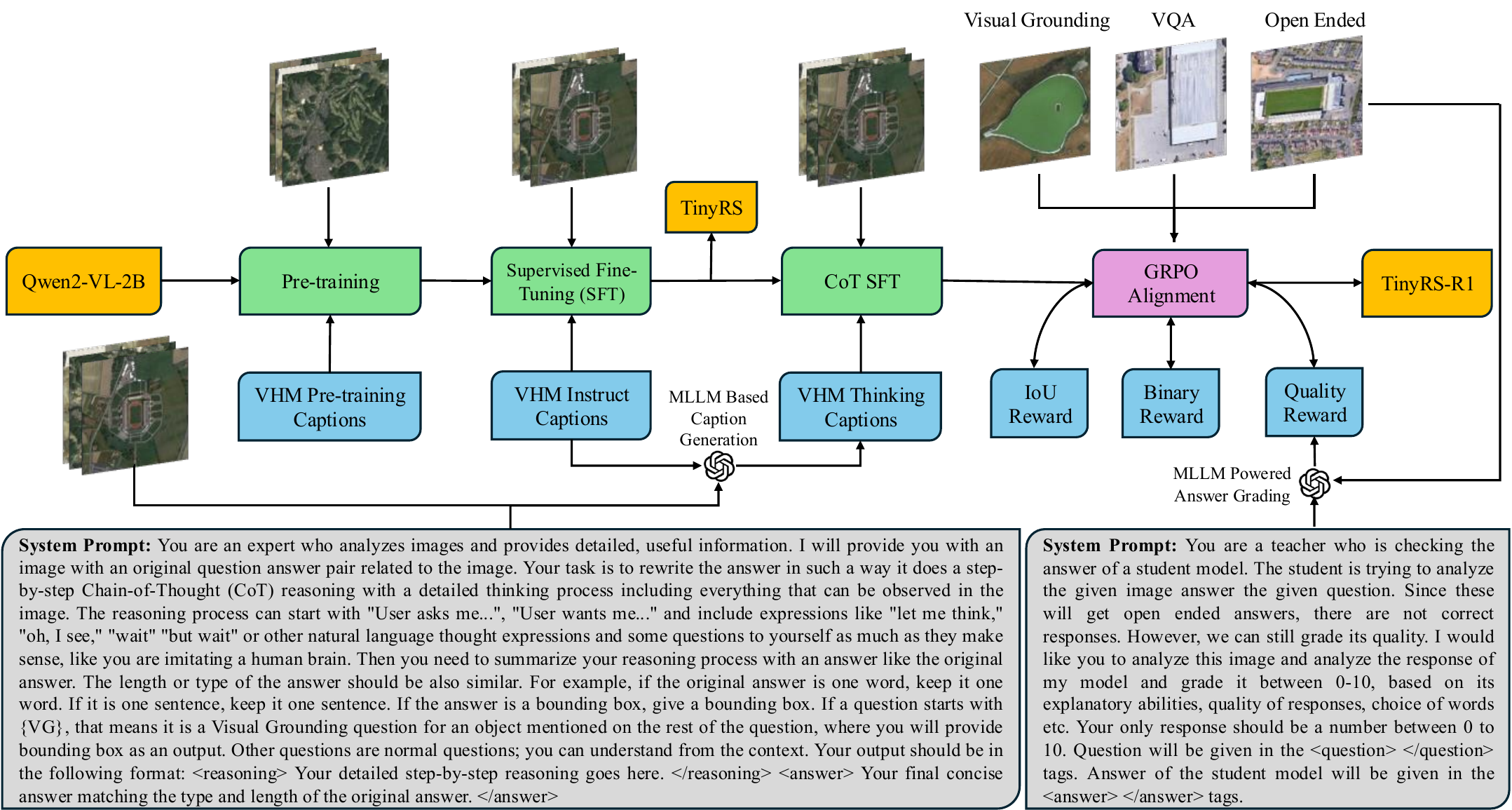}
\end{tabular}
\caption{Training pipeline of \method and \method-R1 involves four stages: VHM pretraining, instruction tuning with VHM-Instruct, CoT fine-tuning via VHM-Instruct-Think, and GRPO-based reward alignment. System prompts for CoT generation and answer grading are shown. \method-R1 includes reasoning and reward feedback; \method is optimized for concise tasks.}
\label{fig:flowchart}
\vspace{-.5cm}
\end{figure*}

Improving accuracy on complex tasks often relies on CoT prompting \cite{nye2021show, wei2022chain}, which guides models through intermediate reasoning steps. To mitigate error propagation, reinforcement learning and step-level supervision have been introduced \cite{uesato2022solving}. These techniques are used in both proprietary models like OpenAI’s o-series \cite{o1preview} and open-source systems like DeepSeek-R1 \cite{guo2025deepseek}, which employ Group Relative Policy Optimization (GRPO) \cite{shao2024deepseekmath} to equip compact multimodal models with strong reasoning abilities.

To address the lack of a small reasoning model for general RS tasks, we introduce \textcolor{revision}{the \method family}. Trained on pre-training and visual instruction datasets \cite{pang2025vhm}, \method matches or outperforms larger models on RS VQA, classification, and visual grounding benchmarks--despite \textcolor{revision}{its smaller size.} Its GRPO-enhanced variant, \method-R1, further boosts performance, achieving state-of-the-art results. To equip \method with reasoning capabilities, we introduce visual reasoning instructions \textcolor{revision}{extending} the VHM visual instruction dataset.

The contributions of this paper are as follows:

\begin{itemize}
    \item We introduce \method, to our knowledge, the smallest open source general purpose RS SVLM, powered by Qwen2-VL.
    \item We propose VHM-Instruct-Think, visual reasoning captions for VHM dataset, which is, to our knowledge, first for RS literature. 
    \item Fine-tuned with reasoning captions then aligned with GRPO, we introduce \method-R1, the first reasoning based SVLM for solving general RS tasks. 
    \item Our experiments show that \method and \method-R1 match or surpass state-of-the-art RS VLMs across most metrics, even against much larger models.
\end{itemize}

While our framework leverages established architectures and alignment techniques, its novelty lies in tailoring these approaches to remote sensing: (i) introducing the first reasoning dataset for RS (VHM-Instruct-Think), (ii) releasing the first reasoning-enabled small-scale RS VLM (\method-R1), and (iii) demonstrating through ablation that reinforcement learning with RS-specific reward signals can meaningfully improve RS benchmarks. Together, these contributions provide both practical tools and new directions for the RS VLM community.

\vspace{-.4cm}
\section{Dataset}
\label{sec:dataset}

All training stages use RS pre-training images and VHM visual instruction data \cite{pang2025vhm}. Pre-training employs original captions over the full VHM set. For supervised fine-tuning, we sample \~100K images--over half of VHM’s instruction set--covering all RS tasks. This subset, called VHM-Instruct, is used in all fine-tuning stages.

Although GRPO can enhance reasoning in fine-tuned language models, it is less effective without prior reasoning-based fine-tuning, especially for smaller models \cite{guo2025deepseek, wei2022chain}. Moreover, most VHM visual instruction captions consist of brief, often one-word or single-sentence responses, making them unsuitable for enabling Chain-of-Thought (CoT) reasoning. To address this, we introduce VHM-Instruct-Think, a set of reasoning-oriented captions generated for the VHM-Instruct dataset.

\begin{figure*}
    \vspace{-.5cm}
    \centering
    \includegraphics[width=\linewidth]{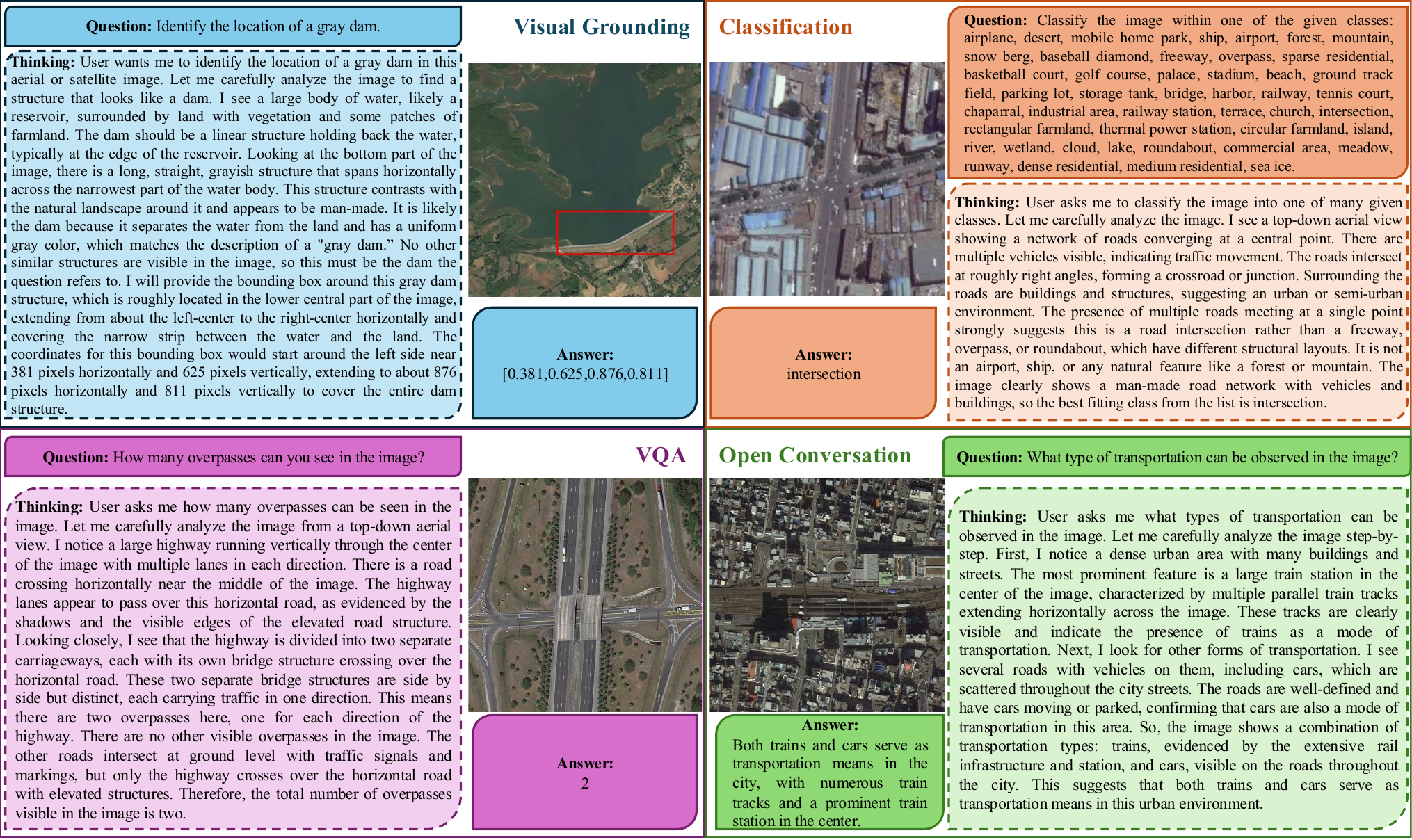}
    \vspace{-.8cm}
    \caption{Examples from the VHM-Instruct-Think dataset, showing satellite images with associated questions, model-generated Chain-of-Thought reasoning, and concise answers. These samples demonstrate the format used for reasoning-augmented supervision in \method-R1.}
    \label{fig:dataset}
    \vspace{-.5cm}
\end{figure*}

Reasoning captions are generated using OpenAI’s GPT-4.1-mini \cite{gpt41}, prompted with the image, VHM’s original question-answer pair, and a system message guiding step-by-step Chain-of-Thought (CoT) reasoning. The prompt encourages natural expressions 
and treats the original answer as metadata. The full prompt appears in Fig. \ref{fig:flowchart}, with sample outputs shown in Fig. \ref{fig:dataset}. This setup yields rich reasoning and concise, answer-aligned outputs for downstream use.

For GRPO training, we use a separate subset of 10K randomly selected images--disjoint from the VHM-Instruct set--sourced from the VHM visual instruction dataset. The question--answer pairs in this subset do not include reasoning, as it is not used for fine-tuning. We refer to this subset as the VHM-RL dataset.

\vspace{-.5cm}
\section{Method}
\label{sec:method}








\noindent TinyRS builds upon Qwen2-VL-2B \cite{wang2024qwen2}, adopting both its architecture and pre-trained weights and follows a 4-step training procedure:

\noindent \textbf{Pre-training and Supervised Fine-Tuning (SFT):} The base \method model is obtained by fine-tuning the original Qwen2-VL-2B-Instruct in two stages. First, it is trained on VHM’s 1M-image pre-training dataset. Then, it is further fine-tuned on the VHM-Instruct dataset. To address task imbalance, underrepresented tasks are upsampled--replicating some examples up to five times per epoch. This two-stage process enables the 2B-parameter model to generalize well across diverse remote sensing tasks.

\noindent \textbf{Chain-of-Thought (CoT) Reasoning:} To improve interpretability and reliability, the model is trained to generate multi-step reasoning that explains visual attributes before giving an answer. This reasoning ability is enabled by fine-tuning the base \method on the VHM-Instruct-Think dataset using CoT prompting. As before, sampling weights are adjusted to balance underrepresented tasks. This equips the 2B-parameter SVLM with structured reasoning and prepares it for GRPO-based reinforcement learning.

\noindent \textbf{Reinforcement Learning:} To further align outputs with classification goals, a lightweight reinforcement learning (RL) phase is applied using Group Relative Policy Optimization (GRPO), which updates the policy based on relative rewards across response groups. Inspired by DeepSeek-R1 \cite{guo2025deepseek} and adapted from \cite{openr1multimodal}, two reward types are used: format-based and accuracy-based.

The format-based reward is binary and granted if the model’s output follows the required structure: \textit{<reasoning>...</reasoning> <answer>...</answer>},
regardless of what is written within the reasoning or final answer.

The accuracy-based reward is task-dependent:

\begin{itemize}
    \item For closed-ended tasks (e.g., VQA, classification, multiple-choice), a binary reward (1 for correct, 0 for incorrect) is used.
    \item For visual grounding, which requires a bounding box as an output, the Intersection over Union (IoU) score is used as the reward.
    \item For open-ended tasks, GPT-4.1-mini serves as an automatic evaluator, scoring the model’s answer from 0 to 10. This score is then normalized to a [0.0, 1.0] range to provide a quality reward. The exact system prompt for the scoring can be found in Fig. \ref{fig:flowchart}.
\end{itemize}

The policy is optimized using GRPO, chosen over standard PPO for its efficiency and better performance on reasoning tasks. GRPO reduces the number of RL steps and helps prevent model collapse or forgetting prior knowledge \cite{shao2024deepseekmath, guo2025deepseek}.

After GRPO training on the VHM-RL dataset, the four-stage pipeline yields the final \method-R1 model. Alternatively, GRPO can be applied directly to the base model--without fine-tuning--using only reward signals. This \textit{zero} approach, inspired by DeepSeek, is evaluated in the ablation study.

\vspace{-.3cm}
\section{Experiments}

\vspace{-.2cm}
\subsection{Settings}

The training pipeline is implemented in PyTorch using HuggingFace Transformers with Qwen2-VL pretrained weights. All stages--pre-training, supervised fine-tuning, and CoT fine-tuning--are run on HPC clusters with 4 $\times$ NVIDIA H100 GPUs, enabling full-parameter tuning of the 2B model. Training is performed for one epoch using Adam optimizer with a learning rate of $1 \times 10^{-5}$ and batch size 16.

The reinforcement learning phase uses GRPO via the TRL framework in batched mode, executed on 16 $\times$ H100 GPUs. GRPO training also runs for one epoch with Adam optimizer (learning rate $1 \times 10^{-6}$), batch size 16, and 4 samples per image. Prompt and completion lengths are extended to 8192 tokens.

\vspace{-.5cm}
\subsection{Results}

\begin{table*}[htbp]
\vspace{-.8cm}
\centering
\small
\caption{Performance of \method on public RS benchmarks compared to previous models. The highest score in each benchmark is marked in \textbf{bold}, and the second is \underline{underlined}.}
\begin{threeparttable}
\setlength{\tabcolsep}{1.8mm}{
\begin{tabular}{llll|lllll} \toprule
    \multirow[t]{3}{*}{\textbf{Capability}} & \multirow[t]{3}{*}{\textbf{Benchmark}} & \textbf{\method} & \textbf{\method} & \textbf{VHM}  & \textbf{Qwen2-VL}  & \textbf{Qwen2-VL} & \textbf{LHRS-Bot} & \textbf{GeoChat} \\
    & & \textbf{R1} &  & \cite{pang2025vhm} & \textbf{RS-R1} \cite{muhtar2025quality} & \textbf{RS} \cite{muhtar2025quality} & \textbf{Nova} \cite{li2024lhrs} & \cite{kuckreja2024geochat}  \\
    & & {\scriptsize thinking} & {\scriptsize non-thinking} & {\scriptsize non-thinking} & {\scriptsize thinking} & {\scriptsize non-thinking} & {\scriptsize non-thinking} & {\scriptsize non-thinking} \\
    & & {\scriptsize 2B} & {\scriptsize 2B} & {\scriptsize 7B} & {\scriptsize 7B} & {\scriptsize 7B} & {\scriptsize 7B} & {\scriptsize 7B} \\    
    \midrule
    \multirow{6}{*}{\shortstack[l]{{\scriptsize Remote}\\{\scriptsize Sensing}\\{\scriptsize Classification}}} & AID & \underline{90.2} & 89.6 & \textbf{92.0} & 82.0 & 84.6 & 83.1 & 73.5 \\
    & METER-ML & 72.1 & 65.1 & \textbf{74.3} & 69.4 & 72.2 & \underline{72.7} & 34.8 \\
    & NWPU & \underline{92.9} & 92.0 & \textbf{94.8} & 84.0 & 86.7 & 83.9 & 89.4 \\
    & SIRI-WHU & \textbf{76.8} & 66.9 & 70.6 & 70.3 & \underline{74.6} & 72.3 & 53.1 \\
    & WHU-RS19 & 95.6 & 91.5 & \textbf{96.5} & 90.6 & 95.5 & \underline{96.2} & 85.5 \\
    \cmidrule{2-9}
    & Average & \textbf{85.6} & 81.0 & \textbf{85.6} & 79.3 & 82.7 & 81.6 & 67.3 \\
    \midrule
    \multirow{6}{*}{\shortstack[l]{{\scriptsize Remote}\\{\scriptsize Sensing}\\{\scriptsize Visual}\\{\scriptsize Question}\\{\scriptsize Answering}}} & HR-Compare & 73.5 & 80.6 & 83.4 & 77.5 & 80.3 & \textbf{89.1} & \underline{83.5} \\
    & HR-Presence & \underline{68.6} & 64.5 & 62.6 & 64.0 & 66.3 & \textbf{84.0} & 57.3 \\
    & LR-Compare & 84.0 & 89.9 & \underline{90.3} & 82.8 & 89.1 & 88.2 & \textbf{90.7} \\
    & LR-Presence & 78.1 & \textbf{90.4} & 89.9 & 74.1 & 86.4 & 84.6 & \underline{90.2} \\
    & LR-Rural & 76.0 & \underline{92.0} & 89.0 & 74.0 & 76.0 & 68.0 & \textbf{96.0} \\  
    \cmidrule{2-9}
    & Average & 76.0 & \textbf{83.5} & 83.0 & 74.5 & 79.6 & 82.8 & \textbf{83.5} \\
    \midrule
    \multirow{1}{*}{\shortstack[l]{{\scriptsize RS Grounding}}} & DIOR-RSVG & \textbf{74.9} & \underline{69.4} & 55.9 & 64.6 & 59.2 & 31.5 & 19.7 \\
    \midrule
    \multirow{1}{*}{\shortstack[l]{{\scriptsize RS Gen. Know.}}} & LHRS-Bench & 56.8 & 57.4 & 33.0 & \textbf{69.2} & \underline{66.5} & 52.5 & 36.2 \\
    \bottomrule
\end{tabular}}
\end{threeparttable}
\vspace{-.5cm}
\label{tab:eval}
\end{table*}

Table~\ref{tab:eval} compares the proposed \method family (2B parameters) with five recent 7B--scale remote--sensing VLMs on four widely known benchmarks: scene classification, visual question answering (VQA), grounding and general RS knowledge.

Most important key observations of Table \ref{tab:eval} are as follows:

\begin{itemize}

\item In classification, \method‐R1 achieves the highest average accuracy (85.6\%), matching VHM and surpassing other 7B baselines by 2-18\%. It leads on SIRI‐WHU and ranks second on AID and WHU‐RS19, showing that lightweight CoT reasoning captures subtle context efficiently. Base \method also performs well (81.0\%), highlighting the benefit of explicit reasoning for fine-grained tasks.

\item For VQA, the trend reverses: the base \method achieves the highest accuracy (83.5\%), matching GeoChat and surpassing all other models. Its concise, single-sentence answers align well with exact-match scoring. By contrast, \method‐R1 drops to 76.0\%, which we attribute to over-elaborated reasoning in binary tasks (e.g., yes/no or urban/rural question answering). In such cases, the additional reasoning steps may occasionally introduce noise, leading to incorrect answers, whereas this effect is not observed in more complex tasks. 

\item On DIOR-RSVG \cite{zhan2023rsvg} grounding, \method‐R1 achieves 74.9\% precision--10\% above the best 7B model and 5.5\% over base \method. 
This demonstrates that CoT reasoning with GRPO enhances performance on complex tasks such as spatial alignment and mitigates the short-answer bias observed in VQA. We attribute this to the IoU reward providing a strong and mathematically precise signal, allowing RL to guide localization more effectively than in linguistic tasks.

\item Both 2B variants lag behind Qwen2-VL-RS on LHRS-Bench \cite{muhtar2024lhrs} but still outperform other baselines. \method slightly surpasses its CoT counterpart (57.4 vs. 56.8\%), indicating that reasoning alone cannot compensate for the limited general knowledge coverage of the VHM training set. This highlights the importance of focusing on LHRS-Bench, demonstrated by the gains of Qwen2-VL-RS-R1.

\end{itemize}

\vspace{-.4cm}
\section{Ablation Study}

In Table \ref{tab:ablation1}, each training stage are turned off at least three times and their effect is reported. 

\noindent \textbf{Effect of remote-sensing pre-training (P).} Comparing with and without pre-training (Line 4 vs. 5) shows modest but consistent gains in scene classification, grounding, and general knowledge, with no change in VQA. Pre-training helps the model grasp aerial features without adding overhead.

\noindent \textbf{Effect of supervised instruction fine-tuning (S).} Turning SFT on/off (Line 2 vs. 6) shows it’s the key performance driver, significantly improving classification, VQA, grounding, and knowledge tasks. It teaches the model what answers to produce, laying the foundation for reasoning and RL.

\noindent \textbf{Effect of Chain-of-Thought fine-tuning (C).} With vs. without CoT (Line 1 vs. 3) shows a trade-off: CoT improves classification, grounding, and knowledge, but slightly hurts VQA--indicating it helps structured tasks but not binary answers, likely due to minor hallucinations leading to incorrect outputs, as discussed earlier.

\noindent \textbf{Effect of GRPO reinforcement learning (R).} Adding GRPO (Line 1 vs. 2) improves grounding significantly and classification modestly, but lowers VQA. 
General knowledge remains unchanged, suggesting RL mainly sharpens spatial precision, as will be further examined with "zero" GRPO.

\noindent \textbf{Effect of “zero” GRPO without P, CoT or S.} GRPO-only (Line 7 vs. 8) boosts grounding by 9.4\% but harms all other tasks. It overfocuses on IoU, producing lower-quality reasoning and repetitive answers--confirming GRPO is only effective after SFT and CoT. This further supports our hypothesis that GRPO is most effective when the reward function (IoU in this case) directly aligns with the evaluation metric.

\begin{table}[bp]
    \centering
    \vspace{-.8cm}
    \caption{Ablation study on training baselines. Average scores are provided for CLS and VQA. P: Pre-training, S: SFT, C: CoT SFT, R: GRPO RL}
    \begin{tabular}{cccccccc}
    \toprule
        \textbf{P} & \textbf{S} & \textbf{C} & \textbf{R} & \textbf{CLS} & \textbf{VQA} & \textbf{Ground.} & \textbf{Gen. Know.} \\
        \midrule
        \checkmark & \checkmark & \checkmark & \checkmark & \textbf{85.6} & 76.0  & \textbf{74.9}  & 56.8 \\
        \checkmark & \checkmark & \checkmark & $\times$ & 84.7 & 78.3  & 70.0 & 57.1 \\
        \checkmark & \checkmark & $\times$  & \checkmark & 82.1 & 78.4 & 69.5  & 55.7 \\
        \checkmark & \checkmark  & $\times$ & $\times$ & 81.0 & \textbf{83.5}  & 69.4  & \textbf{57.4} \\
        $\times$ & \checkmark  & $\times$ & $\times$ & 80.0 & 83.3 & 67.7 & 53.6 \\
        \checkmark & $\times$ & \checkmark & $\times$ & 71.8 & 67.7 & 53.4 & 46.7 \\
        $\times$ & $\times$  & $\times$ & \checkmark & 58.7 & 61.7 & 16.1 & 52.3 \\
        $\times$ & $\times$  & $\times$ & $\times$ & 59.8 & 62.9 & 6.7 & 55.2 \\
    \bottomrule
    \end{tabular}
    \label{tab:ablation1}
    \vspace{-.2cm}
\end{table}

\begin{table}[htbp]
    \centering
    \caption{Ablation study on reward functions of \method-R1. Average scores are provided for CLS and VQA.}
    \begin{tabular}{cccccccc}
    \toprule
        \textbf{Function} & \textbf{CLS} & \textbf{VQA} & \textbf{Ground.} & \textbf{Gen. Know.} \\
        \midrule
        Continuous & 85.6 & \textbf{76.0}  & \textbf{74.9}  & 56.8 \\
        Quantized & 85.3 & 74.1  & 73.6 & 57.3 \\
        Binary & \textbf{85.9} & 75.5 & 72.5  & \textbf{57.7} \\
    \bottomrule
    \end{tabular}
    \label{tab:ablation2}
    \vspace{-.2cm}
\end{table}

\noindent \textbf{Effect of Accuracy Reward Functions in GRPO.} This study evaluates alternatives to the standard continuous rewards used in open-ended and visual grounding tasks. Two variants are tested: \textbf{(1) Binary reward:} assigns 1 if IoU $\geq$ 0.5 (grounding) or GPT-4.1-mini score $\geq$ 0.7 (open-ended), and 0 otherwise. \textbf{(2) Quantized reward:} assigns 1 for IoU/score $\geq$ 0.7, a scaled reward for 0.4-0.7, and 0 below 0.4. As shown in Table \ref{tab:ablation2}, continuous rewards perform best overall, though not always per benchmark. Quantized rewards give stable, balanced results while binary rewards may work for task-specific models like classification or general knowledge but remain suboptimal for general-purpose remote sensing.

\noindent \textbf{Effect of Dataset Balancing.} To assess the impact of dataset balancing--replicating underrepresented tasks up to five times per epoch during SFT--we repeated experiments with the original unbalanced dataset. As shown in Table \ref{tab:ablation3}, balancing significantly boosts classification and VQA performance for both \method and \method-R1, with a slight drop in grounding and general knowledge. Overall, it improves general model performance.

\begin{table}[htbp]
    \centering
    \caption{Ablation study on balancing dataset for \method and \method-R1. Average scores are provided for CLS and VQA. }
    \vspace{-.2cm}
    \begin{tabular}{ccccccccc}
    \toprule
        \textbf{Model} & \textbf{Balanced} & \textbf{CLS} & \textbf{VQA} & \textbf{Ground.} & \textbf{Gen. Know.} \\
        \midrule
        \method & \checkmark & \textbf{81.0} & \textbf{83.5}  & 69.4  & 57.4 \\
        \method & $\times$ & 80.5 & 80.8  & \textbf{70.0} & \textbf{58.1} \\
        \midrule
        \method-R1 & \checkmark & \textbf{85.6} & \textbf{76.0}  & 74.9  & 56.8 \\
        \method-R1 & $\times$ & 80.8 & 73.2 & \textbf{75.1}  & \textbf{57.2} \\
    \bottomrule
    \end{tabular}
    \label{tab:ablation3}
\end{table}

\noindent \textbf{Effect of Model Size.} To assess the 2B model’s effectiveness, we replicated our training strategy with Qwen2-VL-7B. As shown in Table \ref{tab:ablation4}, the 7B model shows slight gains in classification and VQA but underperforms in grounding and general knowledge. This suggests that the dataset is insufficient for stable generalization at the 7B scale, whereas the 2B model is better matched to the available data, making it not only more efficient but also more effective in practice. Memory and speed comparisons further support our claims: 2B models use significantly less memory and run 2-3$\times$ faster than 7B models--both with and without reasoning--making them more suitable for edge deployment.

\begin{table}[htbp]
    \vspace{-.4cm}
    \centering
    \caption{Ablation study on model size with average CLS and VQA scores. \checkmark under Tr. indicates pretraining and fine-tuning with \method parameters; \checkmark under Rs. includes additional CoT and GRPO tuning with \method-R1 parameters. Spd. shows avg. inference time per image on a single H100 (in ms), and Mem. reports avg. gpu memory usage (in GB) in bf16 precision.}
    \scalebox{0.95}{
    \begin{tabular}{cccccccccccc}
    \toprule
        \textbf{Size} & \textbf{Tr.} & \textbf{Rs.} & \textbf{Spd.} & \textbf{Mem.} & \textbf{CLS} & \textbf{VQA} & \textbf{Grnd.} & \textbf{Gen.Knw.} \\
        \midrule
        2B & $\times$ & $\times$ & 90 & 4.4 & 59.8 & 62.9 & 6.7 & 55.2 \\
        2B & \checkmark & $\times$ & 107 & 4.4 & 81.0 & \textbf{83.5} & 69.4 & \textbf{57.4} \\
        2B & \checkmark & \checkmark & 689 & 4.6 & \textbf{85.6} & 76.0 & \textbf{74.9} & 56.8 \\
        \midrule
        7B & $\times$ & $\times$ & 216 & 16.6 & 70.2 & 68.7 & 11.7 & \textbf{64.9} \\
        7B & \checkmark & $\times$ & 232 & 16.6 & 80.8 & \textbf{84.5} & \textbf{38.3} & 62.6 \\
        7B & \checkmark & \checkmark & 1990 & 16.8 & \textbf{86.1} & 80.4 & 31.4 & 64.8 \\
        \bottomrule
    \end{tabular}}
    \label{tab:ablation4}
    \vspace{-.1cm}
\end{table}

\noindent \textbf{Effect of Evaluator LLM.} To address potential evaluator bias, we also experimented with an independently trained evaluator, Haiku 3 \cite{haiku3}. As shown in Table \ref{tab:ablation5}, the overall trends remain consistent, confirming that our results are not dependent on using GPT-4.1-mini as both generator and evaluator.

\begin{table}[htbp]
    \vspace{-.3cm}
    \centering
    \caption{Ablation study on reward functions of \method-R1. Average scores are provided for CLS and VQA.}
    \vspace{-.2cm}
    \begin{tabular}{cccccccc}
    \toprule
        \textbf{Evaluator} & \textbf{CLS} & \textbf{VQA} & \textbf{Ground.} & \textbf{Gen. Know.} \\
        \midrule
        GPT-4.1-mini & \textbf{85.6} & 76.0  & \textbf{74.9}  & 56.8 \\
        Claude Haiku 3 & 84.7 & \textbf{78.0}  & 71.3 & \textbf{59.6} \\
    \bottomrule
    \end{tabular}
    \label{tab:ablation5}
\end{table}

\vspace{-.6cm}
\section{Conclusion}
\label{sec:conclusion}

This paper introduces \method and \method-R1, the first 2B-parameter VLMs for remote sensing, offering a lightweight alternative to 7B-scale models. With domain-specific pretraining, instruction tuning, CoT reasoning, and GRPO alignment, \method-R1 matches or surpasses larger models in classification, grounding, and reasoning. The base \method excels in low-latency VQA, highlighting a trade-off between efficiency and reasoning depth. Both models lag in LHRS-Bench, underscoring the need for general knowledge focus. As a new direction, future work may explore conditional reasoning strategies, such as Mixture-of-Experts approaches, where a router dynamically selects between TinyRS and TinyRS-R1 depending on query complexity.

\vspace{-.2cm}
\section*{Acknowledgements}

The numerical calculations reported in this paper were partially performed using the
MareNostrum 5 pre-exascale supercomputing system and TÜBİTAK TRUBA resources. We gratefully acknowledge the Barcelona
Supercomputing Center (BSC) and the Scientific and Technological Research Council of Turkey
(TÜBİTAK) for providing access to these resources and supporting this research.

\vspace{-.4cm}
\bibliographystyle{IEEEtran}
\bibliography{egbib}

\begin{thebibliography}{10}
\providecommand{\url}[1]{#1}
\csname url@samestyle\endcsname
\providecommand{\newblock}{\relax}
\providecommand{\bibinfo}[2]{#2}
\providecommand{\BIBentrySTDinterwordspacing}{\spaceskip=0pt\relax}
\providecommand{\BIBentryALTinterwordstretchfactor}{4}
\providecommand{\BIBentryALTinterwordspacing}{\spaceskip=\fontdimen2\font plus
\BIBentryALTinterwordstretchfactor\fontdimen3\font minus \fontdimen4\font\relax}
\providecommand{\BIBforeignlanguage}[2]{{%
\expandafter\ifx\csname l@#1\endcsname\relax
\typeout{** WARNING: IEEEtran.bst: No hyphenation pattern has been}%
\typeout{** loaded for the language `#1'. Using the pattern for}%
\typeout{** the default language instead.}%
\else
\language=\csname l@#1\endcsname
\fi
#2}}
\providecommand{\BIBdecl}{\relax}
\BIBdecl

\bibitem{gpt4v}
\BIBentryALTinterwordspacing
OpenAI, ``Gpt-4v(ision) system card,'' 2023. [Online]. Available: \url{https://cdn.openai.com/papers/GPTV_System_Card.pdf}
\BIBentrySTDinterwordspacing

\bibitem{wang2024qwen2}
P.~Wang, S.~Bai, S.~Tan, S.~Wang, Z.~Fan, J.~Bai, K.~Chen, X.~Liu, J.~Wang, W.~Ge \emph{et~al.}, ``Qwen2-vl: Enhancing vision-language model's perception of the world at any resolution,'' \emph{arXiv preprint arXiv:2409.12191}, 2024.

\bibitem{chen2024internvl}
Z.~Chen, J.~Wu, W.~Wang, W.~Su, G.~Chen, S.~Xing, M.~Zhong, Q.~Zhang, X.~Zhu, L.~Lu \emph{et~al.}, ``Internvl: Scaling up vision foundation models and aligning for generic visual-linguistic tasks,'' in \emph{Proceedings of the IEEE/CVF Conference on Computer Vision and Pattern Recognition}, 2024, pp. 24\,185--24\,198.

\bibitem{liu2024mobilellmoptimizingsubbillionparameter}
\BIBentryALTinterwordspacing
Z.~Liu, C.~Zhao, F.~Iandola, C.~Lai, Y.~Tian, I.~Fedorov, Y.~Xiong, E.~Chang, Y.~Shi, R.~Krishnamoorthi, L.~Lai, and V.~Chandra, ``Mobilellm: Optimizing sub-billion parameter language models for on-device use cases,'' 2024. [Online]. Available: \url{https://arxiv.org/abs/2402.14905}
\BIBentrySTDinterwordspacing

\bibitem{hu2307remote}
Y.~Hu, J.~Yuan, C.~Wen, X.~Lu, and X.~Li, ``A remote sensing vision language model and benchmark. arxiv 2023,'' \emph{arXiv preprint arXiv:2307.15266}, 2023.

\bibitem{kuckreja2024geochat}
K.~Kuckreja, M.~S. Danish, M.~Naseer, A.~Das, S.~Khan, and F.~S. Khan, ``Geochat: Grounded large vision-language model for remote sensing,'' in \emph{Proceedings of the IEEE/CVF Conference on Computer Vision and Pattern Recognition}, 2024, pp. 27\,831--27\,840.

\bibitem{muhtar2024lhrs}
D.~Muhtar, Z.~Li, F.~Gu, X.~Zhang, and P.~Xiao, ``Lhrs-bot: Empowering remote sensing with vgi-enhanced large multimodal language model,'' in \emph{European Conference on Computer Vision}.\hskip 1em plus 0.5em minus 0.4em\relax Springer, 2024, pp. 440--457.

\bibitem{pang2025vhm}
C.~Pang, X.~Weng, J.~Wu, J.~Li, Y.~Liu, J.~Sun, W.~Li, S.~Wang, L.~Feng, G.-S. Xia \emph{et~al.}, ``Vhm: Versatile and honest vision language model for remote sensing image analysis,'' in \emph{Proceedings of the AAAI Conference on Artificial Intelligence}, vol.~39, 2025, pp. 6381--6388.

\bibitem{muhtar2025quality}
D.~Muhtar, E.~Zhang, Z.~Li, F.~Gu, Y.~He, P.~Xiao, and X.~Zhang, ``Quality-driven curation of remote sensing vision-language data via learned scoring models,'' \emph{arXiv preprint arXiv:2503.00743}, 2025.

\bibitem{koksal2025milchat}
A.~Koksal and A.~A. Alatan, ``Milchat: Introducing chain of thought reasoning and grpo to a multimodal small language model for remote sensing,'' \emph{arXiv preprint arXiv:2505.07984}, 2025.

\bibitem{nye2021show}
\BIBentryALTinterwordspacing
M.~Nye, A.~J. Andreassen, G.~Gur-Ari, H.~Michalewski, J.~Austin, D.~Bieber, D.~Dohan, A.~Lewkowycz, M.~Bosma, D.~Luan, C.~Sutton, and A.~Odena, ``Show your work: Scratchpads for intermediate computation with language models,'' 2021. [Online]. Available: \url{https://arxiv.org/abs/2112.00114}
\BIBentrySTDinterwordspacing

\bibitem{wei2022chain}
J.~Wei, X.~Wang, D.~Schuurmans, M.~Bosma, F.~Xia, E.~Chi, Q.~V. Le, D.~Zhou \emph{et~al.}, ``Chain-of-thought prompting elicits reasoning in large language models,'' \emph{Advances in neural information processing systems}, vol.~35, pp. 24\,824--24\,837, 2022.

\bibitem{uesato2022solving}
J.~Uesato, N.~Kushman, R.~Kumar, F.~Song, N.~Siegel, L.~Wang, A.~Creswell, G.~Irving, and I.~Higgins, ``Solving math word problems with process-and outcome-based feedback,'' \emph{arXiv preprint arXiv:2211.14275}, 2022.

\bibitem{o1preview}
\BIBentryALTinterwordspacing
OpenAI, ``Introducing openai o1-preview,'' 2024. [Online]. Available: \url{https://openai.com/index/introducing-openai-o1-preview/}
\BIBentrySTDinterwordspacing

\bibitem{guo2025deepseek}
D.~Guo, D.~Yang, H.~Zhang, J.~Song, R.~Zhang, R.~Xu, Q.~Zhu, S.~Ma, P.~Wang, X.~Bi \emph{et~al.}, ``Deepseek-r1: Incentivizing reasoning capability in llms via reinforcement learning,'' \emph{arXiv preprint arXiv:2501.12948}, 2025.

\bibitem{shao2024deepseekmath}
Z.~Shao, P.~Wang, Q.~Zhu, R.~Xu, J.~Song, X.~Bi, H.~Zhang, M.~Zhang, Y.~Li, Y.~Wu \emph{et~al.}, ``Deepseekmath: Pushing the limits of mathematical reasoning in open language models,'' \emph{arXiv preprint arXiv:2402.03300}, 2024.

\bibitem{gpt41}
\BIBentryALTinterwordspacing
OpenAI, ``Introducing gpt-4.1 in the api,'' 2025. [Online]. Available: \url{https://openai.com/index/gpt-4-1/}
\BIBentrySTDinterwordspacing

\bibitem{openr1multimodal}
LMMs-Lab, ``Open-r1 multimodal,'' https://github.com/EvolvingLMMs-Lab/open-r1-multimodal, 2025.

\bibitem{li2024lhrs}
Z.~Li, D.~Muhtar, F.~Gu, X.~Zhang, P.~Xiao, G.~He, and X.~Zhu, ``Lhrs-bot-nova: Improved multimodal large language model for remote sensing vision-language interpretation,'' \emph{arXiv preprint arXiv:2411.09301}, 2024.

\bibitem{zhan2023rsvg}
Y.~Zhan, Z.~Xiong, and Y.~Yuan, ``Rsvg: Exploring data and models for visual grounding on remote sensing data,'' \emph{IEEE Transactions on Geoscience and Remote Sensing}, vol.~61, pp. 1--13, 2023.

\bibitem{haiku3}
\BIBentryALTinterwordspacing
Anthropic, ``Claude 3 haiku: our fastest model yet,'' 2024. [Online]. Available: \url{https://www.anthropic.com/news/claude-3-haiku/}
\BIBentrySTDinterwordspacing

\end{thebibliography}

\end{document}